\documentclass[10pt]{article}

\usepackage[letterpaper, margin=2.54cm, top=2.54cm, bottom=2.54cm]{geometry}
\usepackage[super,comma,sort&compress]{natbib}
\usepackage{lmodern}
\usepackage{authblk} 
\usepackage{amssymb,amsmath}
\usepackage{wrapfig}
\usepackage{graphicx,grffile}
\usepackage[labelfont=bf,labelsep=period]{caption}
\usepackage{ifxetex,ifluatex}
\usepackage{fixltx2e} 
\usepackage{makecell}
\usepackage{multirow}
\usepackage{caption}
\usepackage{subcaption}
\usepackage{array}
\usepackage{wrapfig}
\usepackage{float}
\floatstyle{plaintop}
\restylefloat{table}
\usepackage{dblfloatfix}

\newcolumntype{L}[1]{>{\raggedright\let\newline\\\arraybackslash\hspace{0pt}}m{#1}}
\newcolumntype{C}[1]{>{\centering\let\newline\\\arraybackslash\hspace{0pt}}m{#1}}
\newcolumntype{R}[1]{>{\raggedleft\let\newline\\\arraybackslash\hspace{0pt}}m{#1}}
\ifnum 0\ifxetex 1\fi\ifluatex 1\fi=0 
  \usepackage[T1]{fontenc}
  \usepackage[utf8]{inputenc}
\else 
  \ifxetex
    \usepackage{mathspec}
  \else
    \usepackage{fontspec}
  \fi
  \defaultfontfeatures{Ligatures=TeX,Scale=MatchLowercase}
\fi
\IfFileExists{upquote.sty}{\usepackage{upquote}}{}
\IfFileExists{microtype.sty}{%
	\usepackage{microtype}
	\UseMicrotypeSet[protrusion]{basicmath} 
}{}

\usepackage{sectsty}
\allsectionsfont{\fontsize{10}{10}\selectfont}
\subsectionfont{\normalfont\bfseries\itshape\fontsize{10}{10}\selectfont}
\subsubsectionfont{\normalfont\itshape}

\makeatletter
\renewcommand\subsubsection{\@startsection{subsubsection}{3}{\z@}%
	{-3.25ex\@plus -1ex \@minus -.2ex}%
    {-1.5ex \@plus -.2ex}
    {\normalfont\itshape}}
\makeatother

\makeatletter 
\renewcommand\@biblabel[1]{#1.} 
\makeatother %

\usepackage{etoolbox}
\makeatletter
\patchcmd{\@maketitle}{\LARGE}{\bfseries\fontsize{14}{15}\selectfont}{}{}
\makeatother

\makeatletter
\def\maxwidth{\ifdim\Gin@nat@width>\linewidth\linewidth\else\Gin@nat@width\fi}
\def\maxheight{\ifdim\Gin@nat@height>\textheight\textheight\else\Gin@nat@height\fi}
\makeatother

\setkeys{Gin}{width=\maxwidth,height=\maxheight,keepaspectratio}
\ifx\paragraph\undefined\else
\let\oldparagraph\paragraph
\renewcommand{\paragraph}[1]{\oldparagraph{#1}\mbox{}}
\fi
\ifx\subparagraph\undefined\else
\let\oldsubparagraph\subparagraph
\renewcommand{\subparagraph}[1]{\oldsubparagraph{#1}\mbox{}}
\fi
\usepackage{hyperref}
\hypersetup{
	unicode=true,
	pdftitle={Natural Language Processing Methods to Identify Oncology Patients at High Risk for Acute Care with Clinical Notes},
	pdfauthor={Claudio Fanconi},
	pdfkeywords={Natural Language Processing, Machine learning and predictive modeling},
	breaklinks=true
}
\title{\vspace{-2em} Natural Language Processing Methods to Identify Oncology Patients at High Risk for Acute Care with Clinical Notes}
\author[ ]{\bf\fontsize{12}{13}\selectfont Claudio Fanconi, B.Sc.\textsuperscript{1,2}, Marieke van Buchem,
M.Sc.\textsuperscript{1,3}, Tina Hernandez-Boussard, Ph.D., M.P.H., M.Sc.\textsuperscript{1}\vspace{-.7em}}
\affil[1]{\bf\fontsize{12}{13}\selectfont Stanford University, Stanford, California, United States}
\affil[2]{\bf\fontsize{12}{13}\selectfont ETH Zürich, Zürich, Switzerland}
\date{} 
\affil[3]{\bf\fontsize{12}{13}\selectfont Leiden University Medical Center, Leiden, The Netherlands}

\begin{document}
\maketitle
\vspace{-4em} 

\section{Abstract}\label{abstract}
\emph{Clinical notes are an essential component of a health record. This paper evaluates how natural language processing (NLP) can be used to identify the risk of acute care use (ACU) in oncology patients, once chemotherapy starts. Risk prediction using structured health data (SHD) is now standard, but predictions using free-text formats are complex. This paper explores the use of free-text notes for the prediction of ACU in leu of SHD. Deep Learning models were compared to manually engineered language features. Results show that SHD models minimally outperform NLP models; an $\ell_1$-penalised logistic regression with SHD achieved a C-statistic of 0.748 (95\%-CI: 0.735, 0.762), while the same model with language features achieved 0.730 (95\%-CI: 0.717, 0.745) and a transformer-based model achieved 0.702 (95\%-CI: 0.688, 0.717). This paper shows how language models can be used in clinical applications and underlines how risk bias is different for diverse patient groups, even using only free-text data.}

\section{Introduction}\label{introduction}
Oncology patients undergoing chemotherapy often need acute care utilisation (ACU) and hospitalisation after starting chemotherapy. These interventions account for nearly half of the costs associated with oncology care in the United States \cite{cost_1, cost_2}. Evidence suggests roughly 50\% of these healthcare encounters are potentially preventable through early outpatient interventions\cite{preventable_1, preventable_2}. A previous paper by Peterson et al~\cite{dylan_paper} introduced a machine learning (ML) model, using structured health data (SHD) from electronic health records (EHR), to identify patients at high risk for ACU after chemotherapy initiation. In total, they trained their model using 760 inputs and retained 125 to predict the risk of ACU. This work, and others, highlight the potential of data-driven models to predict ACU risk using SHD\cite{keiserboys1, canadaboy, Daly2019AFF}.\\
However, most EHRs are not mapped to a common data model and they are not necessarily standardised between different facilities. To replicate other hospital’s predictive models they could require intensive data preparation. On the other hand, 96\% of hospitals in the US collect digital clinical notes from physicians and nurses in 2019~\cite{ehr}. Natural language processing (NLP) methods can extract useful information from these unstructured clinical texts.\\
NLP methods have already proven useful in clinical applications, e.g. for predicting critical care outcomes in intensive care units~\cite{ben_paper}, classifying procedures and diagnoses~\cite{ben_2}, or predicting outcomes after an ischemic stroke~\cite{stroke}. In particular, deep learning-based language models have become popular in recent years~\cite{nlp_review}, being used to identify 30-day hospital readmissions~\cite{clinicalbert, clinicalbert_2} or Statin non-use~\cite{tina_1}.\\
This study aims to assess the added value for identifying patients at risk of needing ACU by replacing tabular inputs with features from unstructured clinical notes or by combining both modalities. Second, we aim to investigate whether novel deep learning language models outperform traditional language feature extraction and linear models. We investigated these aspects by developing five predictive models of ACU risk trained with different inputs and compared their predictive performance and utility when used at the point of care.

\section{Methods}\label{methods}
\subsection{Data Collection}\label{data_collection}
In 2019, the Centers for Medicare \& Medicaid Services (CMS) introduced the Chemotherapy Measure (also referred to as OP-35), a quality measure that captures hospital admissions or emergency department visits of adult patients related to potentially preventable diagnoses within 30 days of starting outpatient chemotherapy \cite{op_35}. Based on this measure, a study population at a comprehensive cancer center, including a large tertiary outpatient clinic, was assembled for risk prediction at 30, 180 and 365 days after chemotherapy initiation~\cite{dylan_paper}. The OP-35 metric itself is used as a label for supervised learning and defines a positive event.\\
For the SHD inputs, we use the original 760 features from Peterson et al.~\cite{dylan_paper} extracted from the same EHR database, such as demographic, social, vital sign, procedural, diagnostic, medication, laboratory, health care utilisation, and cancer-specific data generated 180 days before the first date of chemotherapy. For a detailed description of how the patient cohort was extracted, the inclusion and exclusion criteria for the OP-35 metric, and a full list of features, please see the original paper~\cite{dylan_paper}.\\
Based on the above study population, we matched patients to their respective progress notes and the history and physical (H\&P) notes from the EHR database (Epic Systems Corp). We removed notes of less than 100 words, as these were mainly erroneous entries, and notes of more than 5,000 words, as these often contained long copies of previous notes and laboratory analyses. We also removed history notes with mentions of clinical trial consents, as based on our review these were copy-paste texts. Finally, we extract and aggregate the most recent clinical notes (at most three) created 180 days before the patient started chemotherapy, same as in the SHD collection from Peterson et al.~\cite{dylan_paper}. If a patient had no clinical records in the EHR database, they were removed from the study population.\\
The cohort was previously randomly divided into a training set (80\%) and a test set (20\%) for modelling, and we, therefore, kept exactly these patient sets (except the ones without any clinical notes) to obtain comparable results.\\

\subsection{Model Development}
\begin{figure*}[]
  \centering
  \includegraphics[width=\textwidth]{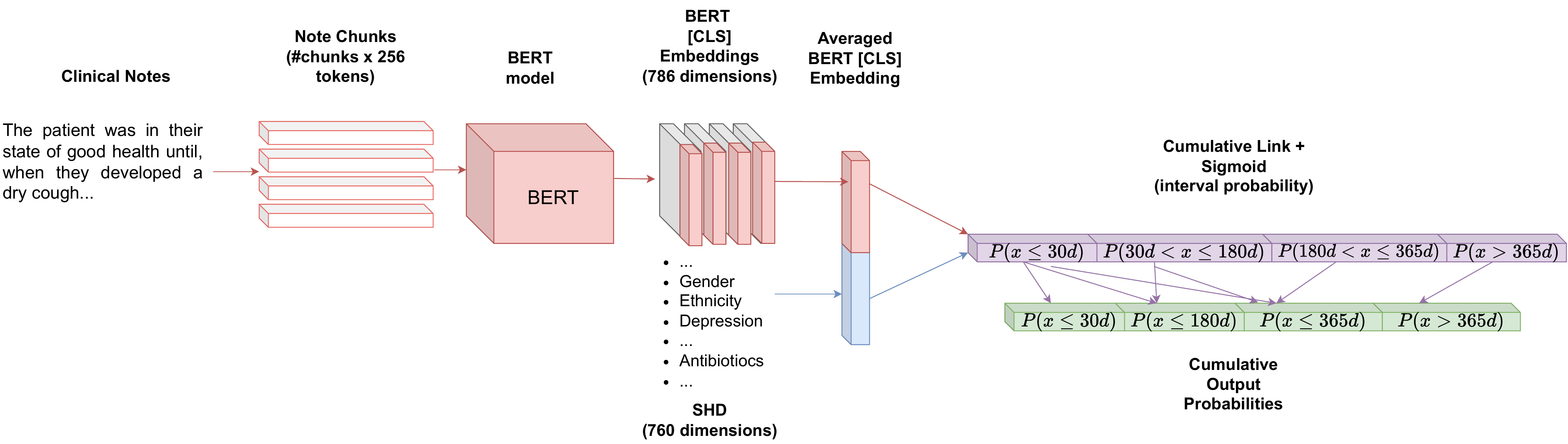}
  \caption{Overview of the Fusion BERT model. The Language BERT only contains the upper (red) encoder architecture before leading into the output probabilities (purple and green).}
  \label{fig:fusion_bert}
\end{figure*}

Five different risk prediction models were compared in this study: \textit{Tabular LASSO}, \textit{Language LASSO}, \textit{Fusion LASSO}, \textit{Language BERT} and \textit{Fusion BERT}.\\
\textbf{Tabular LASSO.} This model is a logistic regression with an $\ell_1$-penalty (also known as Least Absolute Shrinkage and Selection Operator - LASSO). The inputs were the 760 structured health data points from Peterson et al.~\cite{dylan_paper}.\\
\textbf{Language LASSO.} The Language LASSO model is an $\ell_1$-penalised logistic LASSO regression with manually generated inputs from the clinical notes. The notes were preprocessed as follows: First, we removed special characters and personal, organisational, date and time entities using SpaCy's~\cite{medspacy} part of speech tagging. Then we tagged negated terms with a "not\_" using SpaCy's negator library. We removed auxiliary words, adpositions, determiners, interjections and pronouns. Subsequently, we lemmatised~\cite{wordnet} the remaining words. Finally, we followed the method of Marafino et al.~\cite{ben_paper} by filtering out the 2,000 most frequent terms\footnote{We also tested the algorithm with 500, 1,000, and 3,000 most occurring terms. While increasing the number of terms increased the predictive performance, the increase became insignificant after 2,000. We omitted these results from the main paper because of space constraints.} of all the notes and weighting these words using the Term Frequency-Inverse Document Frequency (TF-IDF) algorithm. The Language LASSO has 2,000 input features corresponding to the 2,000 most frequently occurring words.\\
\textbf{Fusion LASSO.} The Fusion LASSO is also a logistic regression LASSO model. This time it uses both, the tabular data and TF-IDF values, as input features. We combined them to inspect if data extracted from the clinical notes has added value to SHD.\\
\textbf{Language BERT.} This model is a deep learning-based Bidirectional Encoder Representation of Transformers~\cite{transformer, bert} (BERT). This model does not require manual feature engineering and can consume clinical notes with little preprocessing. We used a pre-trained distilBERT~\cite{distilbert} model as the encoding structure, as it requires less computation than the more familiar BERT or ClinicalBERT~\cite{clinicalbert} models\footnote{Using a pre-trained ClinicalBERT over a pre-trained distilBERT did not yield any significant improvements in prediction. We, therefore, omitted its results from the paper due to space constraints.}. DistilBERT and other transformer models are often computationally limited by the input size of the text (often referred to as token length in the deep learning literature). To avoid these GPU memory overflows, we decomposed the clinical notes into chunks of at most 25 sequences, each 256 tokens (1 token $\approx$ 1 word), and ran them through the neural network. We aggregated the outputs of the transformers (also called embeddings) by averaging over the embeddings of the corresponding clinical note. We connected the averaged embedding (representing a complete clinical note) linearly to a single output neuron, whose value is divided into four sections. A sigmoid function is applied to assign a probability to each of these values. These four slices represent the probability distribution of an ACU event within the time intervals emanating from the different ground truth labels ($P(x \le 30d)$, $P(30d < x \le 180d)$, $P(180d < x \le 365d)$ and $P(x > 365d)$). Since a patient who experienced an ACU event within the first 30 days is also eligible for an event within 180 days and 365 days, we add the corresponding probabilities to get the original ground truth interpretation of an ACU within 30 days ($P(x\le 30d)$), 180 days ($P(x \le 180d)$), 365 days ($P(x \le 365d)$) and not within 365 days ($P(x > 365d)$).\\
\textbf{Fusion BERT}. The fusion BERT model is the same as the language BERT model, except that the corresponding SHD are concatenated with the output embedding. The newly-concatenated embedding was then linearly connected to the output neuron. Figure~\ref{fig:fusion_bert} shows an overview of the fusion BERT.\\
The regularisation hyperparameters of the LASSO models were determined by tenfold cross-validation grid search, while the hyperparameters of the two BERT models were determined by using 20\% of the training data as validation data. While the LASSO models are trained on each time interval of the label (30d, 180d, and 365d) individually, the BERT models are trained on all the labels simultaneously. We applied a cumulative link loss~\cite{cll} to train the neural network with backpropagation, to ensure the ordinal regression structure\cite{rosenthal_2018} (e.g avoid cases where $P(x \le 30d) > P(x\le 180d)$, as in this use case any patient that had an ACU event within 30 days, subsequently is marked to have had an ACU within 180).

\subsection{Model Evaluation}
We first evaluated the models by their discriminative performance with the area under the receiver operating characteristic curve (AUROC), the area under the precision-recall curve (AUPRC) and the negative log-likelihood (cross-entropy) with a 1'000-fold bootstrap to obtain 95\% confidence intervals.\\
We reported the number of SHD used during risk prediction. We did this for the LASSO models by summing the number of non-zero model coefficients originating from SHD. For the Fusion BERT, we counted the number of connections of tabular features to the output neuron that have values less than 0.001, as the backpropagation algorithm is not optimized for feature selection, unlike the LASSO.\\
To assess the model calibration, calibration curves~\cite{Van_Calster2016-ml} were developed. Finally, we assessed the initial clinical utility of these four models through a Decision Curve Analysis (DCA)~\cite{net_benefit}. A DCA plots the net benefit across a range of decision thresholds and quantifies the number of true positives versus false positives. The curves for the prediction models were compared to two alternative clinical strategies: treat all (everyone is treated as if they will have an ACU event) and treat none (nobody is treated as if they will have an ACU event).\\
To test the discriminatory power of the model in a setting similar to that in which it might be used at the point of care, the test cohort was stratified into high, medium and low-risk groups based on the tertile of predicted risk. Kaplan-Meier~\cite{kaplan-meyer} survival curves for OP-35 events are used to examine the separation between risk groups for language LASSO and language BERT on 180-day ACU risk prediction. In addition, the ten highest and lowest coefficients of the language LASSO model are presented. This helps us to determine the importance of certain keywords in the clinical notes.\\
Since Peterson et al.~\cite{dylan_paper} have reported unfair algorithmic results for ACU prediction from structured data, we investigated whether language features might also be affected. We present and compare the empirical cumulative distributions of predicted risk score percentiles for subgroups to assess how the models predicted each subgroup's risk for OP-35 events. Specifically, we examine demographic values (i.e. race and insurance type) and tumour stage on the Language LASSO model for 180-day ACU risk prediction.\\

\section{Results}\label{results}
\begin{table*}[h!]
\centering
\scriptsize
\begin{tabular}{l|r|r|r|r|r}
\multicolumn{1}{c|}{\multirow{3}{*}{\textbf{Patient   Characteristic}}} & \textbf{Total Cohort} & \textbf{OP-35 Events} & \textbf{OP-35 Events} & \textbf{OP-35 Events} & \textbf{No OP-35 Events} \\
\multicolumn{1}{c|}{} & ~ & \textbf{within 30 days} & \textbf{within 180 days} & \textbf{within 365 days} & \textbf{within 365 days} \\
\multicolumn{1}{c|}{} & \textbf{(N=6,938)} & \textbf{(n=936, 13.5\%)} & \textbf{(n=2,202, 31.7\%)} & \textbf{(n=2,704, 39.0\%)} & \textbf{(n=4,234, 61.0\%)} \\ \hline\hline
\textbf{Age, mean}$\pm$ \textbf{std} & \textbf{} & \textbf{} & \textbf{} & \textbf{} & \textbf{} \\
At diagnosis & 58.7$\pm$14.3 & 57.2$\pm$15.3 & 57.7$\pm$15.1 & 57.9$\pm$15.0 & 59.2$\pm$13.9 \\
At first chemotherapy & 60.5$\pm$14.4 & 58.9$\pm$15.2 & 59.4$\pm$15.1 & 59.6$\pm$15.0 & 61.0$\pm$14.0 \\ \hline
\textbf{Sex, No. (\%)} & \textbf{} & \textbf{} & \textbf{} & \textbf{} & \textbf{} \\
Female & 3,659 (52.7) & 474 (50.6) & 1132 (51.4) & 1417 (52.4) & 2242 (53.0) \\ \hline
\textbf{Race, No (\%)} & \textbf{} & \textbf{} & \textbf{} & \textbf{} & \textbf{} \\
White & 3,804 (54.8) & 461 (49.3) & 1,113 (50.5) & 1,379 (51.0) & 2,425 (57.3) \\
Asian & 1,619 (23.3) & 226 (24.1) & 536 (24.3) & 649 (24.0) & 970 (22.9) \\
Black & 188 (2.7) & 42 (4.5) & 88 (4.0) & 100 (3.7) & 88 (2.1) \\
Other or unknown & 1,327 (19.1) & 207 (22.1) & 465 (21.1) & 576 (21.3) & 751 (17.7) \\ \hline
\textbf{Ethnicity, No. (\%)} & \textbf{} & \textbf{} & \textbf{} & \textbf{} & \textbf{} \\
Non Hispanic/Latino & 5,989 (86.3) & 788 (84.2) & 1,867 (84.8) & 2,280 (84.3) & 3,709 (87.6) \\
Hispanic or Latino & 855 (12.3) & 142 (15.2) & 327 (14.9) & 414 (15.3) & 441 (10.4) \\ \hline
\textbf{Cancer type, No. (\%)} & \textbf{} & \textbf{} & \textbf{} & \textbf{} & \textbf{} \\
Breast & 1,321 (19.0) & 113 (12.1) & 275 (12.5) & 346 (12.8) & 975 (23.0) \\
Gastrointestinal & 819 (11.8) & 93 (9.9) & 291 (13.2) & 366 (13.5) & 453 (10.7) \\
Thoracic & 774 (11.2) & 107 (11.4) & 258 (11.7) & 326 (12.1) & 448 (10.6) \\
Lymphoma & 700 (10.1) & 170 (18.2) & 345 (15.7) & 382 (14.1) & 318 (7.5) \\
Head and neck & 658 (9.5) & 90 (9.6) & 208 (9.4) & 238 (8.8) & 420 (9.9) \\
Pancreas & 585 (8.4) & 99 (10.6) & 214 (9.7) & 280 (10.4) & 305 (7.2) \\
Prostate & 520 (7.5) & 11 (1.2) & 46 (2.1) & 70 (2.6) & 450 (10.6) \\
Gynecologic & 513 (7.4) & 70 (7.5) & 176 (8.0) & 218 (8.1) & 295 (7.0) \\
Genitourinary & 461 (6.6) & 76 (8.1) & 184 (8.4) & 219 (8.1) & 242 (5.7) \\
Other & 587 (8.5) & 107 (11.4) & 205 (9.3) & 259 (9.6) & 328 (7.7) \\ \hline
\textbf{Cancer stage, No. (\%)} & \textbf{} & \textbf{} & \textbf{} & \textbf{} & \textbf{} \\
Stage I & 1,099 (15.8) & 123 (13.1) & 281 (12.8) & 338 (12.5) & 761 (18.0) \\
Stage II & 1,415 (20.4) & 141 (15.1) & 336 (15.3) & 410 (15.2) & 1005 (23.7) \\
Stage III & 964 (13.9) & 131 (14.0) & 351 (15.9) & 429 (15.9) & 535 (12.6) \\
Stage IV & 1,898 (27.4) & 327 (34.9) & 759 (34.5) & 937 (34.7) & 961 (22.7) \\
Unknown & 1,562 (22.5) & 214 (22.9) & 475 (21.6) & 590 (21.8) & 972 (23.0) \\\hline
\textbf{Insurance, No. (\%)} & \textbf{} & \textbf{} & \textbf{} & \textbf{} & \textbf{} \\ 
Medicare & 2,683 (38.7) & 323 (34.5) & 788 (35.8) & 970 (35.9) & 1,713 (40.5) \\
Private & 2,450 (35.3) & 328 (35.0) & 747 (33.9) & 898 (33.2) & 1,552 (36.7) \\
Medicaid & 599 (8.6) & 130 (13.9) & 258 (11.7) & 307 (11.4) & 292 (6.9) \\
Other or unknown & 1,206 (17.4) & 155 (16.6) & 409 (18.6) & 529 (19.6) & 677 (16.0) \\ \hline
\end{tabular}
    \caption{Information about the complete patient cohort (train and test set) eligible for the OP-35 metric for 30, 180, and 365 day prediction. "std" stands for standard deviation.}\label{tab:patient_cohort}
\end{table*}

A total of 6,938 patients were included in the study cohort compared to the original cohort, which included 8,439 patients. The mean age at chemotherapy initiation was 60.5 years ($\pm$ 14.4 years), and 52.7\% were female. A total of 936 patients (13.5\%) met the primary criteria of having at least one OP-35 event within the first 30 days of starting chemotherapy, 2,202 (31.7\%) within the first 180 days and 2,704 (39.0\%) within the first year. The majority of patients in the cohort were white (n=3,804, 54.8\%), followed by Asian patients (n=1,619, 23.3\%), then other and unknown races (1,327, 19.1\%) and least represented were black patients (n=188, 2.7\%). The most common cancer type was breast cancer (n=1,321, 19.0\%), gastrointestinal tumours (n=819, 11.8\%), thoracic cancer (n=774, 11.2\%) and lymphoma (n=700, 10.1\%), which accounted for more than half of all data. ACU events occurred most frequently in lymphoma (30d: n=170, 18.2\%; 180d: n=345, 15.7\%; 365d: n=382, 14.1\%) and least frequently in prostate cancer (30d: n=11, 1.2\%; 180d: n=46, 2.1\%; 365d: n=70, 2.6\%) across all time periods. Most chemotherapy patients had a stage IV tumour (n=1,898, 27.4\%), which was also most common in ACU events (30d: n=327, 34.9\%; 180d: n=759, 34.5\%; 365d n=937, 34.7\%). The most common type of insurance in the cohort was Medicare (n=2,863, 38.7\%) and private health insurance (n=2,450, 35.3\%). The cohort characteristics are summarised in Table~\ref{tab:patient_cohort}.

\begin{table*}[h]
\centering
\scriptsize
\begin{tabular}{c|l|r|r|r|r}
\multicolumn{1}{l|}{\textbf{Label}} & \textbf{Model} & \textbf{No. SHD} & \textbf{AUROC} & \textbf{AUPRC} & \textbf{Cross-Entropy}\\ \Xhline{3\arrayrulewidth}
\multirow{10}{*}{30} & Tabular LASSO & 83    & 0.775 & \textbf{0.411} & 0.344\\
 & C=0.02 & ~ & (0.757,0.792) & (0.373,0.447) & (0.329,0.358)\\ \cline{2-6} 
 & Language LASSO  & N/A                      & 0.726 & 0.294 & 0.363 \\
 & C=0.03 & ~ & (0.707,0.744) & (0.264,0.323) & (0.346,0.379)\\ \cline{2-6} 
 & Fusion LASSO & 73                         & \textbf{0.778} & 0.410 & \textbf{0.341}\\
 & C=0.02 & ~ & (0.760,0.795) & (0.372,0.447) & (0.326,0.356)\\ \cline{2-6} 
 & Language BERT & N/A                      & 0.710 & 0.259 & 0.435\\
 &  & ~ & (0.692,0.729) & (0.235,0.282) & (0.415,0.455)\\ \cline{2-6} 
 & Fusion BERT & ~ 419                        & 0.766 & 0.315 & 0.393\\
 &  & ~ & (0.749,0.784) & (0.286,0.343) & (0.377,0.406)\\ \hline\hline
\multirow{10}{*}{180} & Tabular LASSO & ~ 221  & 0.748 & 0.623 & 0.540\\
 & C=0.03 & ~ & (0.735,0.762) & (0.600,0.647) & (0.527,0.552)\\ \cline{2-6} 
 & Language LASSO & ~  N/A                     & 0.730 & 0.577 & 0.558\\
 & C=0.02 & ~ & (0.717,0.745) & (0.555,0.601) & (0.546,0.570)\\ \cline{2-6} 
 & Fusion LASSO & ~ 101                       & \textbf{0.765} & \textbf{0.632} & \textbf{0.530}\\
 & C=0.02 & ~ & (0.752,0.779) & (0.610,0.655) & (0.517,0.543)\\ \cline{2-6} 
 & Language BERT & ~ N/A                        & 0.702 & 0.543 & 0.625\\
 &  & ~ & (0.688,0.717) & (0.517,0.567) & (0.603,0.644)\\ \cline{2-6} 
 & Fusion BERT & ~  419                        & 0.753 & 0.620 & 0.548 \\
 &  & ~ & (0.741,0.767) & (0.597,0.644) & (0.536,0.558)\\ \hline\hline
\multirow{10}{*}{365} & Tabular LASSO & ~ 150  & 0.763 & \textbf{0.704} & 0.559\\
 & C=0.02 & ~ & (0.752,0.775) & (0.685,0.724) & (0.549,0.569)\\ \cline{2-6} 
 & Language LASSO & ~ N/A                      & 0.732 & 0.639 &0.585\\
 & C=0.02 & ~ & (0.719,0.745) & (0.618,0.661) & (0.575,0.595)\\ \cline{2-6} 
 & Fusion LASSO & ~ 115                       & \textbf{0.770} & 0.702 & \textbf{0.553}\\
 & C=0.02 & ~ & (0.759,0.782) & (0.683,0.722) & (0.541,0.563)\\ \cline{2-6} 
 & Language BERT & ~ N/A                       & 0.709 & 0.617 & 0.666\\
 &  & ~ & (0.695,0.723) & (0.594,0.640) & (0.647,0.683)\\ \cline{2-6} 
 & Fusion BERT & ~  419                        & 0.760 & 0.695 & 0.565 \\
 &  & ~ & (0.748,0.774) & (0.675,0.714) & (0.554,0.575)\\ \hline
\end{tabular}
\caption{Resulting metrics on the test set of the tabular, language and fusion LASSO models, as well as the language and fusion BERT, trained on 30, 180 and 365 days ACU prediction with the 95\%-CI in the brackets. The best-performing metrics for every label type are marked in bold. We also display the number of SHD used for prediction in the third column, where "N/A" means that SHD was used for prediction.}\label{tab:metrics}
\end{table*}

\subsection{Model Performance}\label{model_performance}
\begin{figure*}[b!]
\begin{subfigure}{.19\textwidth}
  \centering
  \includegraphics[width=\textwidth]{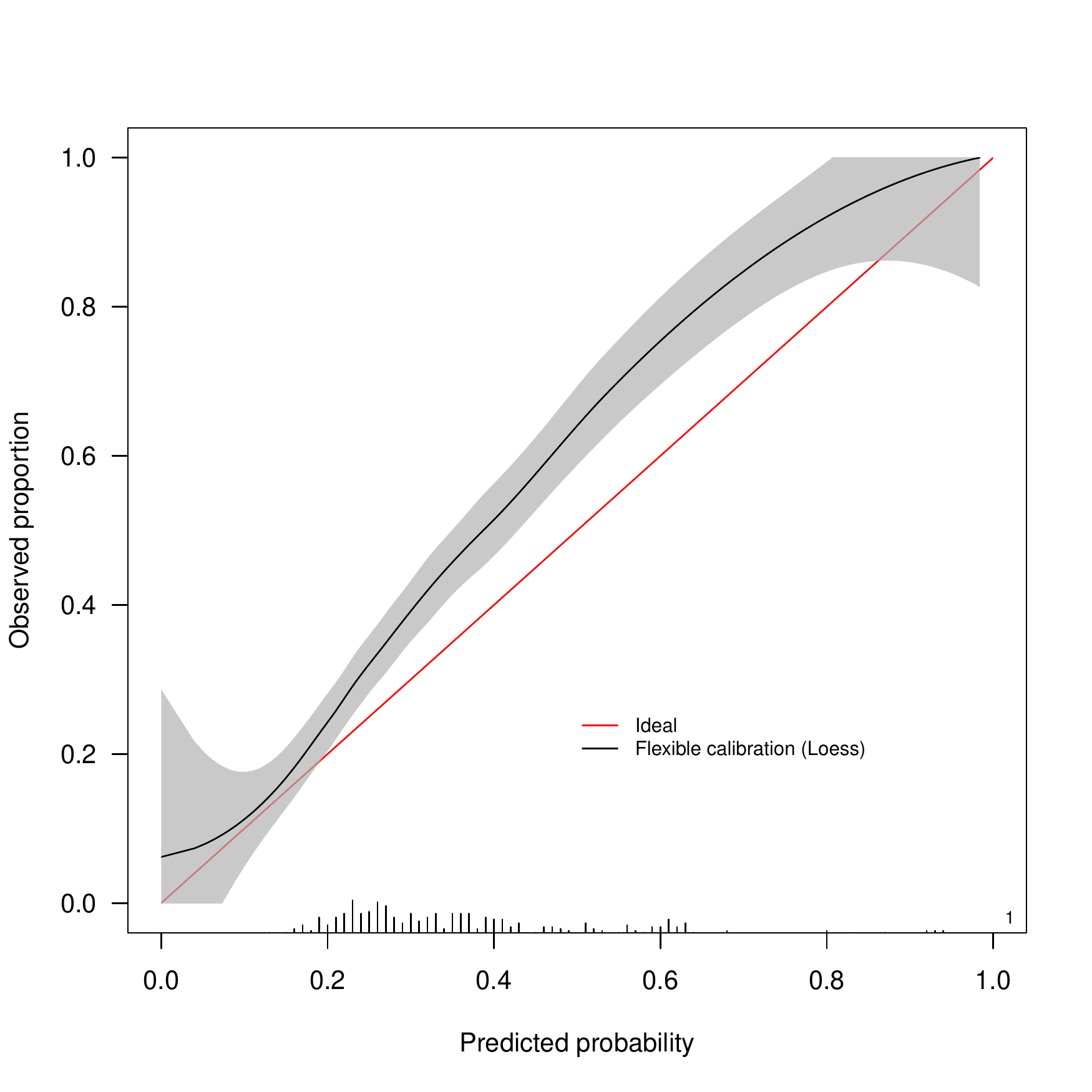}
  \caption{Tabular LASSO}
\end{subfigure}
\begin{subfigure}{.19\textwidth}
  \centering
  \includegraphics[width=\textwidth]{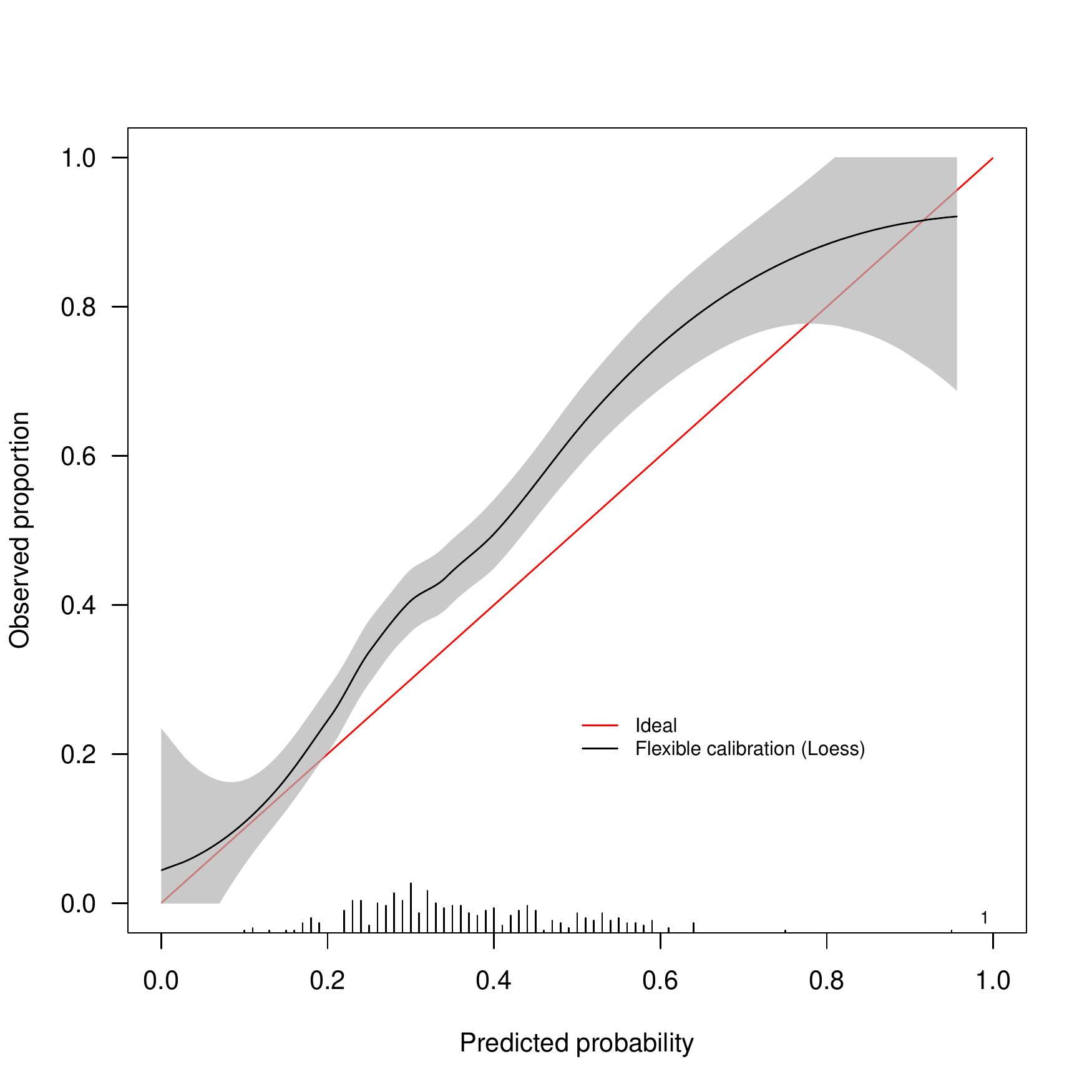}
  \caption{Language LASSO}
  \label{fig:sfig5}
\end{subfigure}
\begin{subfigure}{.19\textwidth}
  \centering
  \includegraphics[width=\textwidth]{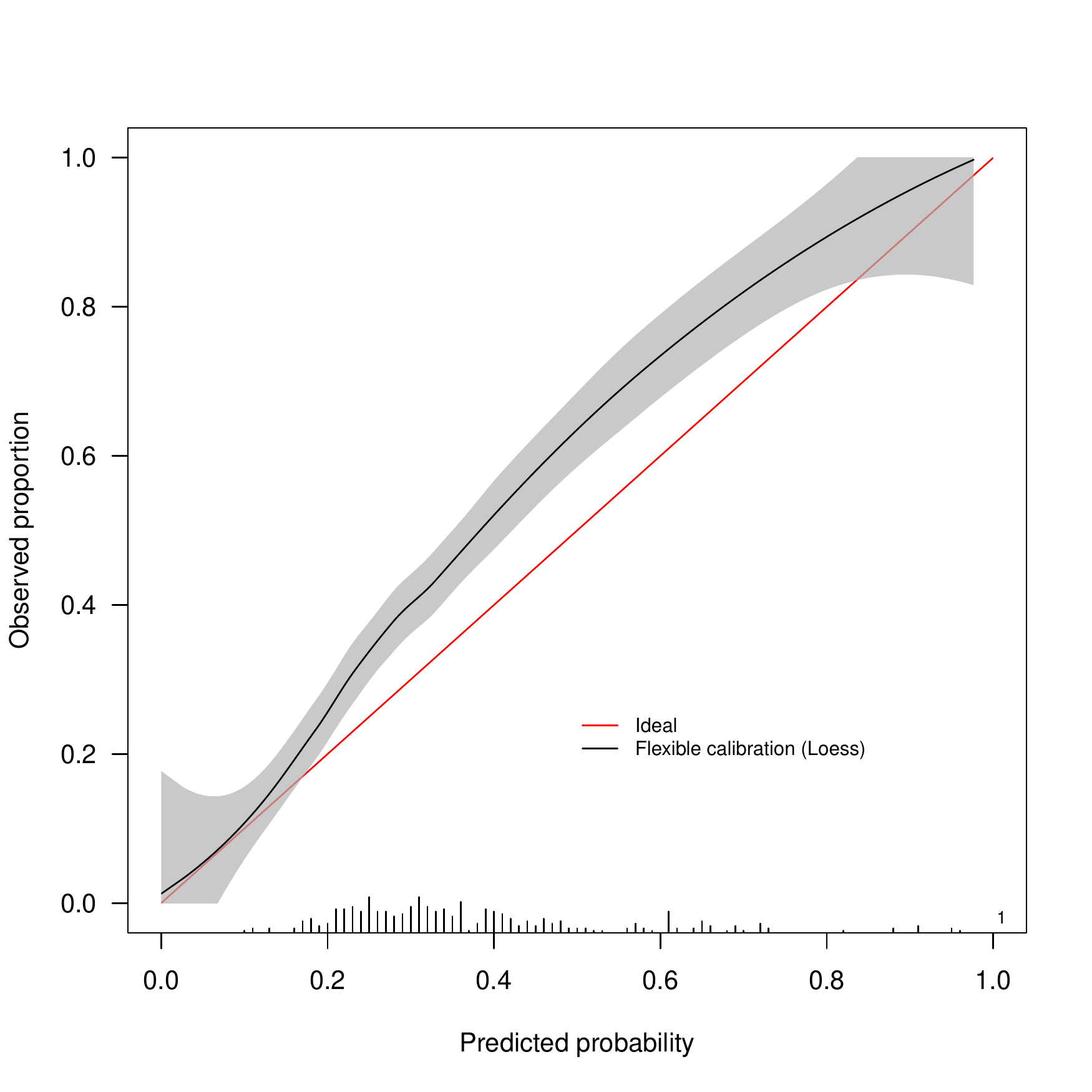}
  \caption{Fusion LASSO}
  \label{fig:sfig8}
\end{subfigure}
\begin{subfigure}{.19\textwidth}
  \centering
  \includegraphics[width=\textwidth]{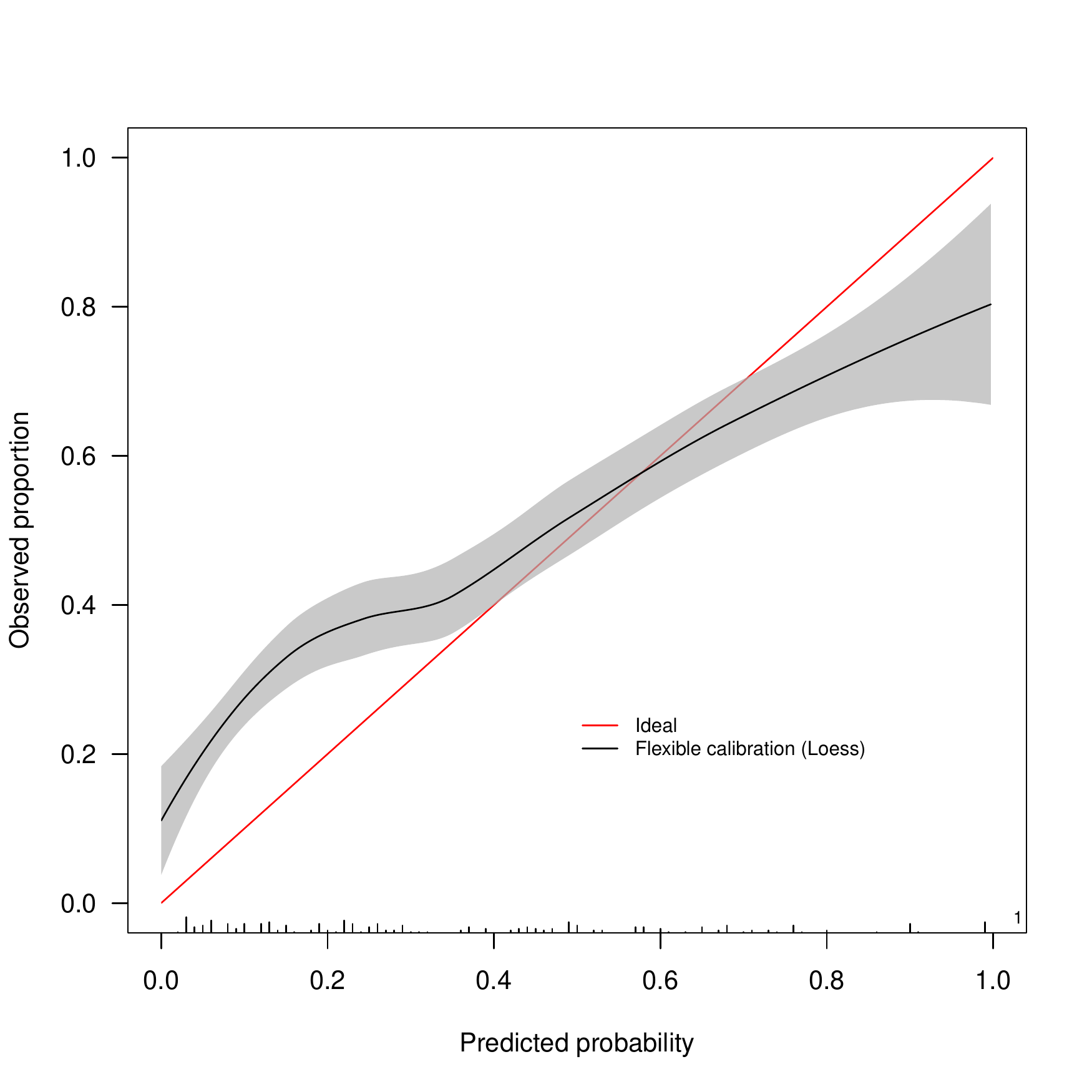}
  \caption{Language BERT}
  \label{fig:sfig11}
\end{subfigure}
\begin{subfigure}{.19\textwidth}
  \centering
  \includegraphics[width=\textwidth]{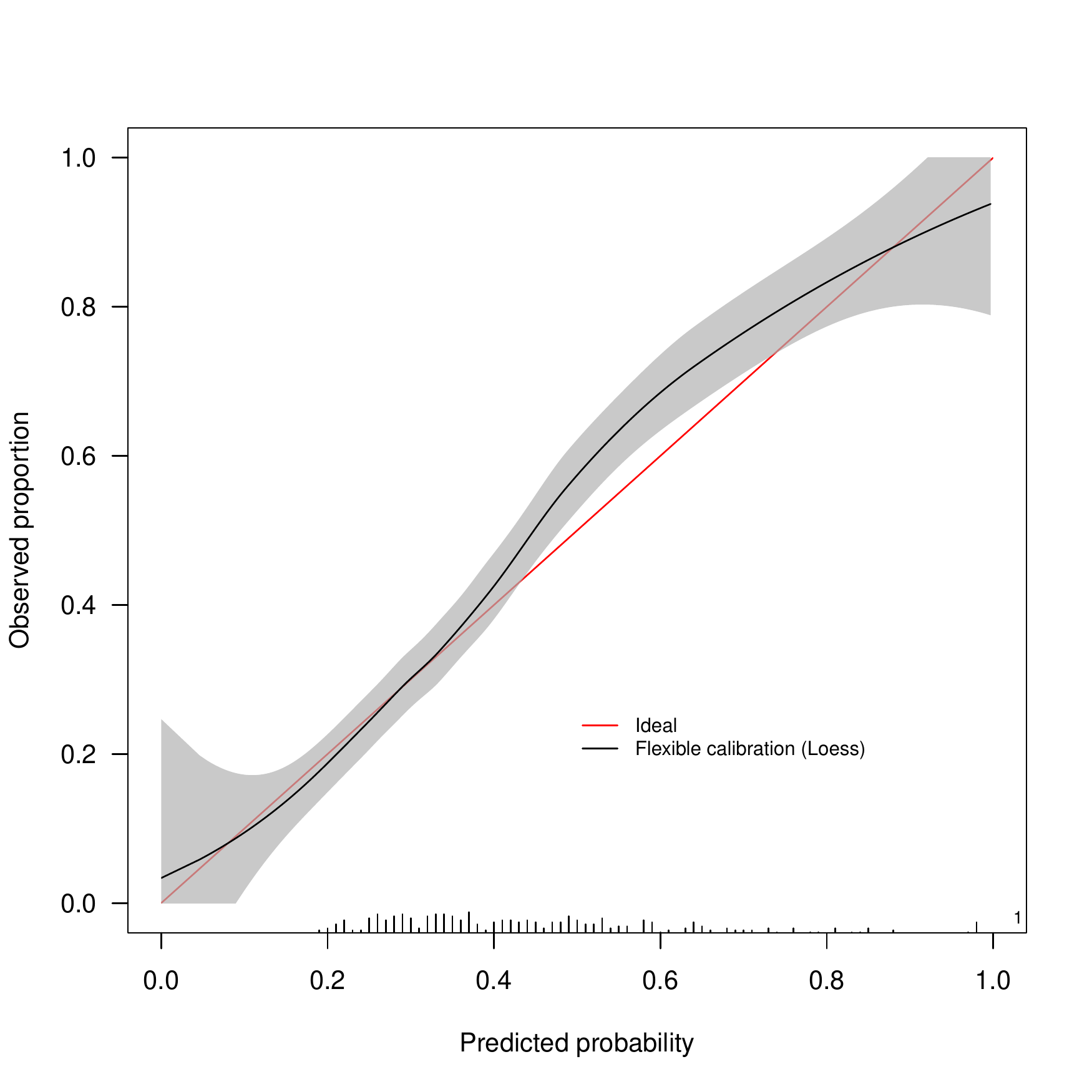}
  \caption{Fusion BERT}
  \label{fig:sfig14}
\end{subfigure}
    \caption{Calibration curves of the 180-day ACU risk prediction models. The red line indicates ideal calibration, while the black line is the flexible calibration with the 95\% confidence interval, generated with the prob.cal.ci.2 function\cite{Van_Calster2016-ml}.}
    \label{fig:calibration}
\end{figure*}
Table~\ref{tab:metrics} lists the AUROC, AUPRC, and cross-entropy scores including the 95\% confidence intervals of the five risk models for 30-day, 180-day and 365-day ACU prediction. For the 30 day acute care risk prediction, the Fusion LASSO model performs best on AUROC (0.778, 95\%-CI: 0.760, 0.795) and cross-entropy (0.341, 95\%-CI: 0.326, 0.356), using 73 SHD features. The highest AUPRC score has the Tabular LASSO (0.411, 95\%-CI: 0.373, 0.447) compared to the event rate of 13.5\%, using 83 tabular variables.\\
For 180-day ACU prediction, the Fusion LASSO model performs best in all metrics with 101 SHD features. The Language LASSO has a 0.730 (95\%-CI: 0.717, 0.745) AUROC score and the Language BERT achieves 0.702 (95\%-CI: 0.688, 0.717), both of them without using any structured data.\\
In the full-year ACU prediction, we observe that the Fusion LASSO scores again the highest C-stastic (0.770, 95\%-CI:0.759, 0.782) and the lowest cross-entropy loss (0.553, 95\%-CI:0.541, 0.563), using 115 tabular features, while the Tabular LASSO has the highest AUPRC score (0.704, 95\%-CI:0.685, 0.724), using 150 SHD points.\\
We show the flexible calibration curves for the 180-day models in Figure~\ref{fig:calibration}, where we observe a risk underestimation of the three LASSO models and underestimation of low risk patients and overestimation of high risk patients with the Language BERT model.\\
The Fusion BERT uses the most SHD points (419 tabular inputs) for all three label types to make predictions.

\subsection{Exploration of Clinical Usage of Language Models}
\begin{figure}[h!]
    \centering
    \includegraphics[width=0.8\textwidth]{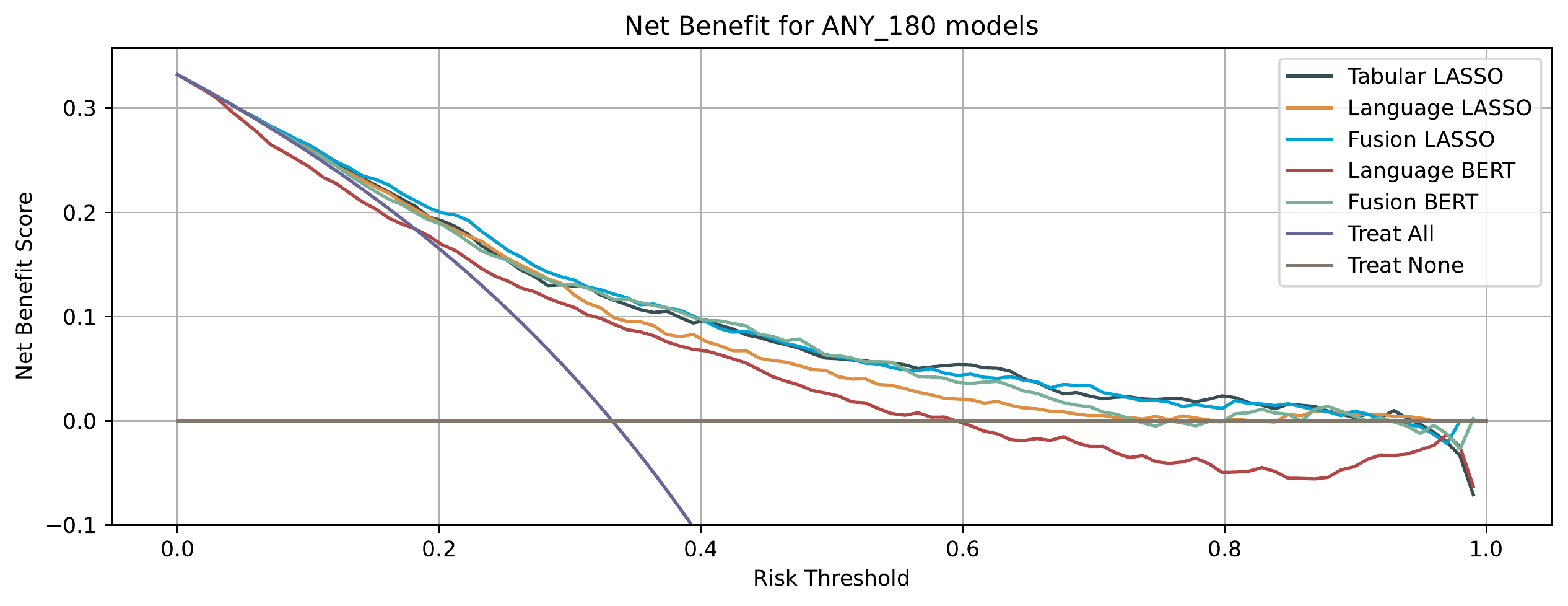}
    \caption{Net benefit curves of the tabular, language and fusion models. The purple curve indicates the benefit of all the patients treated, whereas the grey curve indicates the benefit is no patient is treated}
    \label{fig:net_benefit}
\end{figure}

\begin{figure}[]
\begin{subfigure}{.5\textwidth}
  \centering
  \includegraphics[width=\textwidth]{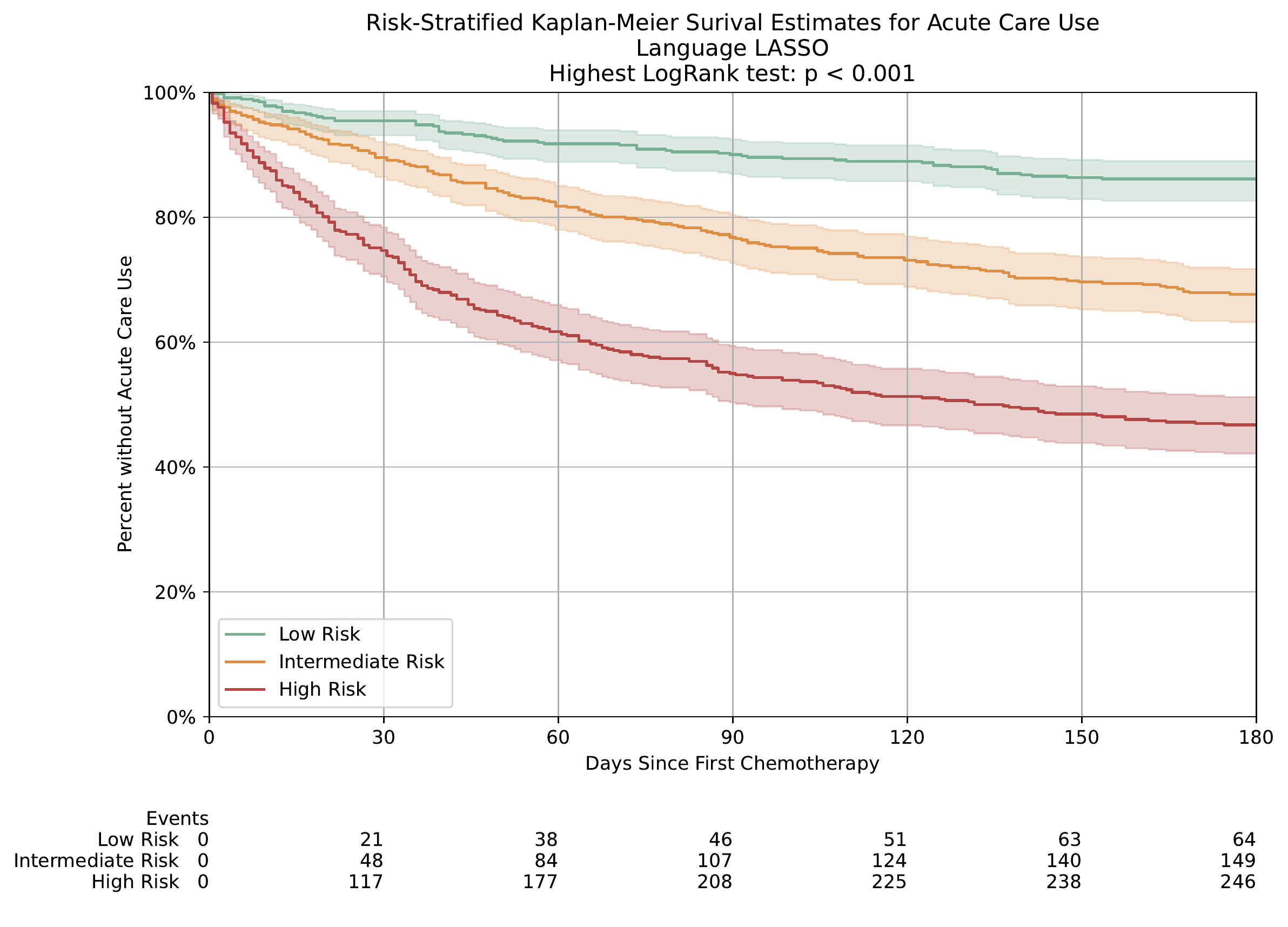}
  \caption{Language LASSO (180d predictions)}
  \label{fig:sfig1}
\end{subfigure}%
\begin{subfigure}{.5\textwidth}
  \centering
  \includegraphics[width=\textwidth]{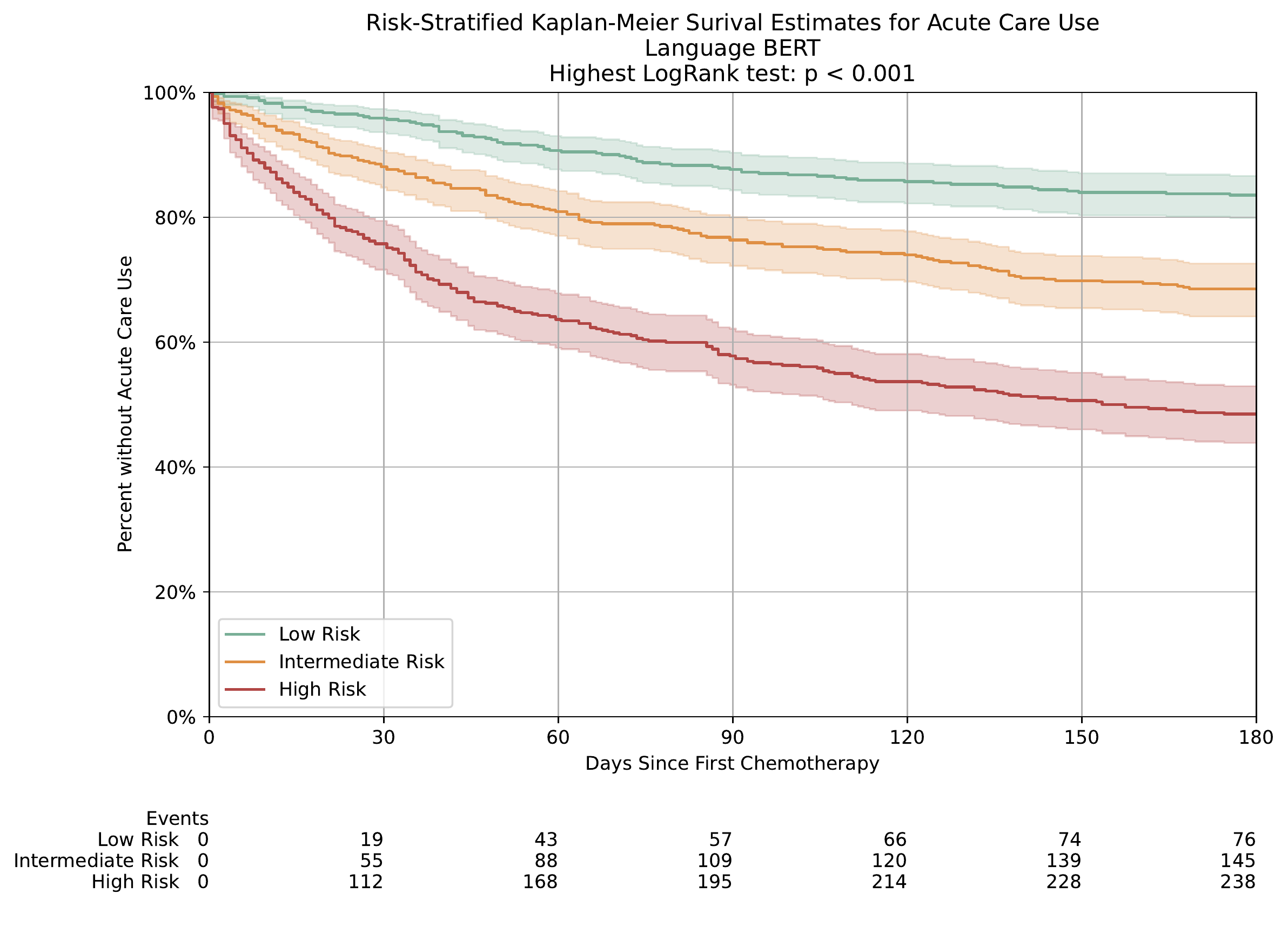}
  \caption{Language BERT (180d predictions)}
  \label{fig:sfig2}
\end{subfigure}
    \caption{Kaplan-Meier curves for ACU events for patients in the test cohort stratified by predicted risk. The shaded area represents the 95\%CIs.}
    \label{fig:kaplan_meier}
\end{figure}

\begin{figure}[h!]
\centering
    \includegraphics[width=0.6\textwidth]{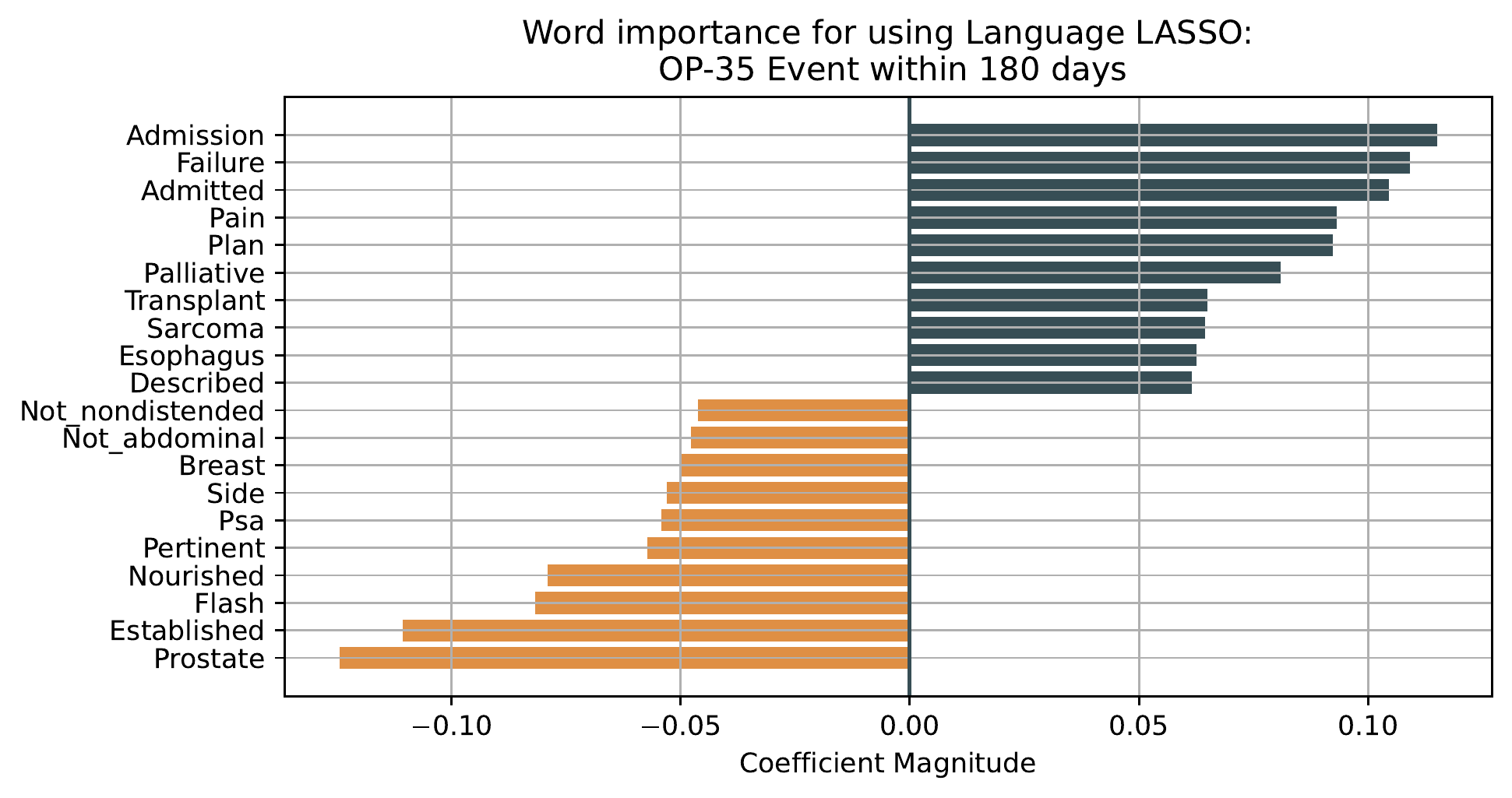}
    \caption{Coefficient magnitudes for the Language LASSO for 180-day ACU prediction, displaying the ten highest and ten lowest. The coefficients in this model are single words found in the clinical notes before the ACU event.}
    \label{fig:word_importance}
\end{figure}

The Decision curve analysis for the 180-day ACU prediction showed that the net benefit of the Language-BERT model yields a negative benefit when the decision threshold for treatment is chosen above 0.6 (Figure~\ref{fig:net_benefit}) and less or equal net benefit than treating every patient with a threshold below 0.19. The other models, including the Language-LASSO model, have positive benefit values for decision thresholds below 0.7.\\
The Kaplan-Meier survival curves for OP-35 events showed good separation between risk groups (Figure~\ref{fig:kaplan_meier}, p < 0.001 for each group by log-rank test) for the two language-only models. By 180 days after the start of chemotherapy, 64 (13.9\%) of the 462 low-risk patients in the language LASSO prediction had an OP-35 event and 76 (16.5\%) in the language BERT prediction. On the other hand, 246 (53.2\%) of the 462 high-risk patients had an event for the speech LASSO prediction and 238 (51.5\%) for the language BERT prediction.\\
Figure~\ref{fig:word_importance} shows the relative importance of the ten highest and lowest coefficients of the language LASSO model for the 180-day prediction. The words "Admission", "Failure", "Pain" and "Palliative" are among the ten highest coefficients, while "Breast", "PSA (Prostate-specific antigen)", "Nourished" and "Prostate" are among the ten lowest coefficients.

\begin{figure}[h!]
\centering
\begin{subfigure}{.33\textwidth}
  \centering
  \includegraphics[width=\textwidth]{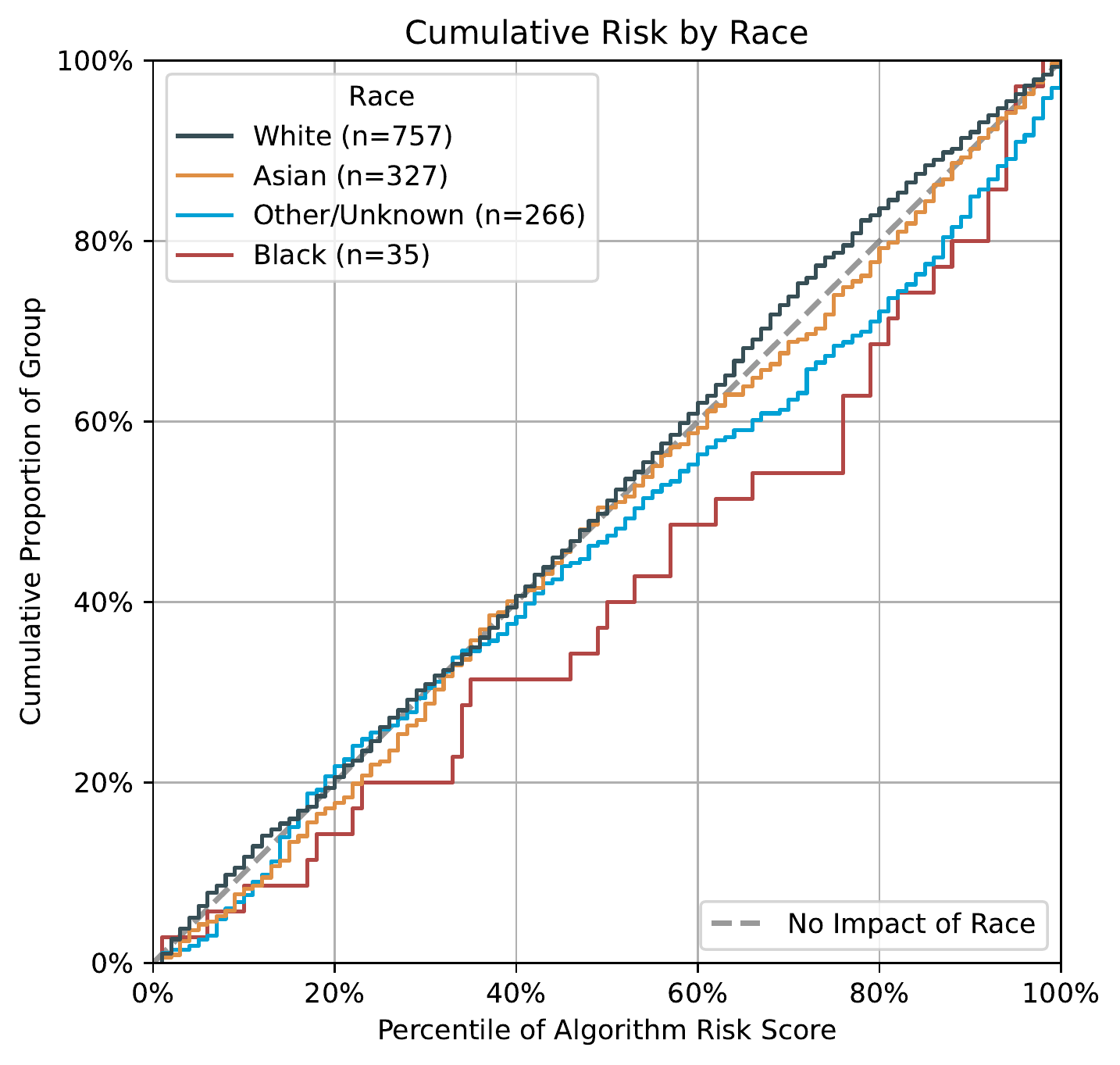}
  \caption{Cumulative risk by race}
  \label{fig:race}
  \end{subfigure}%
\begin{subfigure}{.33\textwidth}
  \centering
  \includegraphics[width=\textwidth]{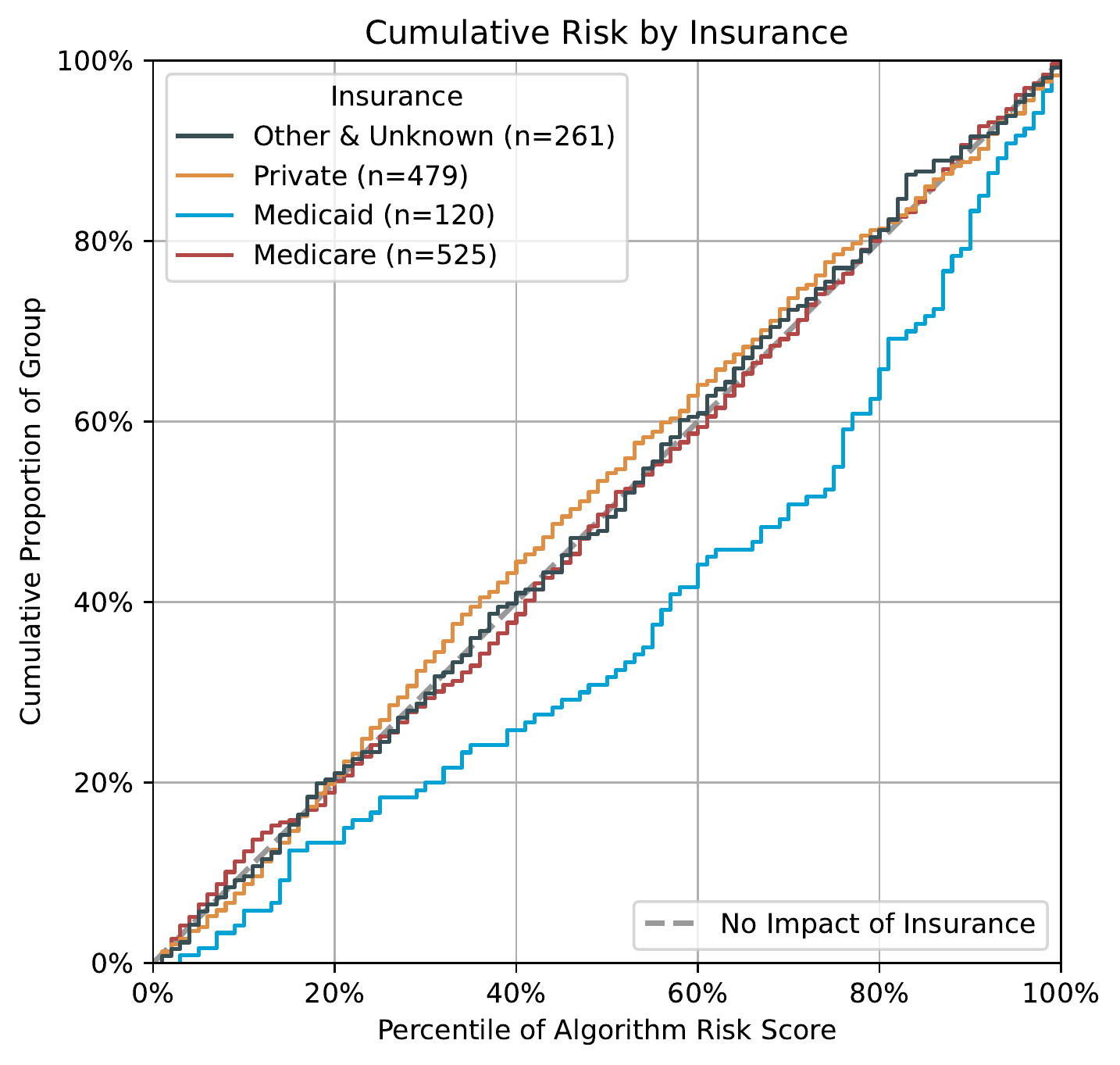}
  \caption{Cumulative risk by insurance}
  \label{fig:insurance}
  \end{subfigure}
\begin{subfigure}{.33\textwidth}
  \centering
  \includegraphics[width=\textwidth]{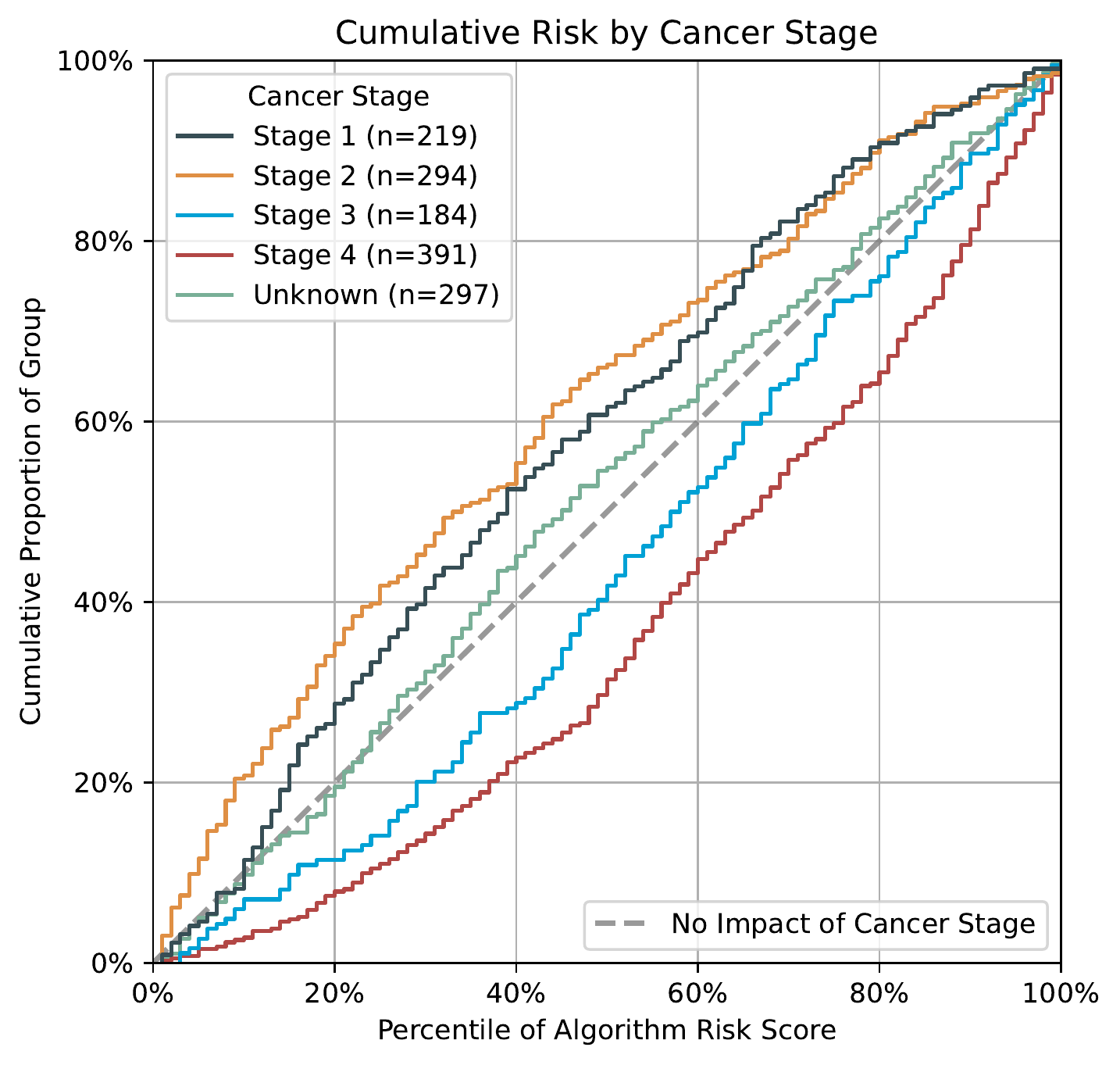}
  \caption{Cumulative risk by cancer stage}
  \label{fig:cancer_type}
  \end{subfigure}
    \caption{Cumulative risk of the Language LASSO on 180 ACU prediction, stratified by race (a), insurance type (b) and over cancer stage (c).}
    \label{fig:sensitivity_analysis}
\end{figure}

\subsection{Sensitivity Analysis}
Figure~\ref{fig:race} shows that black patients are predicted to have a disproportionately higher risk than white, Asian or other-race patients. We note that the number of black patients is at least seven times smaller than that of non-black races. The cumulative risk by different insurance types is displayed in Figure~\ref{fig:insurance}, where we note a risk overestimation of Medicaid patients.\\
In Figure~\ref{fig:cancer_type} we see that the risk predictions for patients with stage III and IV tumours are overestimated, while the predictions for patients with stage I, II and unknown stage cancer are underestimated.

\section{Discussion}

Identifying methods to improve external generalisability is a priority for the medical informatics community. This paper presented an analysis of NLP to identify patients at risk of seeking acute care for different time intervals, a method which is scalable across sites and required minimal data preprocessing. It presents a comparison of logistic regression LASSO using structured health data with manually generated language features and the combination of both modalities. In addition, these models are compared to deep learning-based transformers using either full-text clinical records only or in combination with structured health data.\\
Results demonstrate tabular LASSO outperforming language LASSO and language BERT at all time intervals. Nevertheless, we point out that both language-based methods achieve good discriminative performance (AUROC $\ge 0.7$) even without the use of tabular health data, especially for 180-day risk prediction. On the other hand, combining the two input modalities (clinical notes and SHD) did not yield significant improvements over using tabular data alone.
Regarding the selection model, results show that the popular BERT-based classifier does not outperform $\ell_1$-penalised logistic regression with TF-IDF features and the fusion of both input modalities. This is likely due to the aggregation of chunks of the lengthy clinical documents into a single output probability, which makes deep learning training difficult.\\
Our study has several strengths; first, we show that NLP methods can be used instead of high-dimensional SHD to identify chemotherapy patients at risk for acute treatment. Secondly, our method optimises considerably training as we developed a transformer-based model that is trained simultaneously on multiple risk intervals in the form of nested ordinal regression, so that the computationally intensive training of the model for the different labels is not required.\\ 
Three main clinical implications can be drawn from this study. First, ACU risk prediction models for chemotherapy patients perform well using free-text data from the last (at most three) H\&P and progress notes from physicians before chemotherapy. This implies that NLP methods could be easily implemented across sites and/or facilities as they only require access to written medical notes without the need for re-structuring or mapping of structured data, potentially saving costs in feature collection. Second, we show that LASSO coefficients can be used by clinicians to understand relative word meaning when making a prediction. BERT models lack this type of interpretive mechanism that allows clinicians to build confidence in the model. Third, in the sensitivity analysis, we find that certain groups may be subject to risk bias, despite not using demographic values explicitly as inputs. Although this needs further investigation, the results suggest that  clinicians should be aware of these subgroup differences when interpreting the results of ML models.\\
Our study has limitations. First, more sophisticated transformer language models can handle longer notes~\cite{longformer} or also cross-modality~\cite{cross_modal_attention}. However, these techniques require significant computation, which may not be feasible at most institutes. Second, our models have been validated only on one dataset for risk prediction in acute care. It would be beneficial to test the language models in a variety of medical problems. Finally, the work relies on data from a single academic cancer center, which may be not generalisable to other populations.\\
In summary, this study demonstrates the utility of using free-text data to identify patients at risk of needing acute care once they have started chemotherapy. It is an alternative to structured health data, which may require significant preprocessing and may not be generalisable across settings. We show that the Language LASSO is a suitable model, especially for 180-day prediction, and that it is still interpretable. This work advances our knowledge on risk prediction models and provides an alternative for cross site generalisability. Hospital systems may consider these techniques to validate risk models.

\section{Data and Code Availability}
Under the terms of the data sharing agreement for this study, we cannot share source data directly. Requests for anonymous patient-level data can be made directly to the authors. All experiments were implemented in Python using the library SciKit-learn~\cite{scikit} for the metrics and the classical logistic regression model, and PyTorch~\cite{pytorch} for the transformer models. We used R to create the calibration plots. The code for the logistic LASSOs and our analysis is available on \href{https://github.com/su-boussard-lab/nlp-for-acu}{www.github.com/su-boussard-lab/nlp-for-acu}, while the code for the BERT model and training is available on \href{https://github.com/su-boussard-lab/bert-for-acu}{www.github.com/su-boussard-lab/bert-for-acu}.

\bibliographystyle{vancouver}
\bibliography{main}

\end{document}